\documentclass[letterpaper, 10 pt, conference]{ieeeconf}  

\IEEEoverridecommandlockouts                              

\usepackage{graphicx} 
\usepackage{xcolor}
\usepackage{amsmath} 
\usepackage{amssymb}  
\usepackage{cite} 
\usepackage{multirow} 
\usepackage{url}
\title{\LARGE \bf
Autonomous UAV–Quadruped Docking in Complex Terrains via Active Posture Alignment and Constraint-Aware Control}

\author{Haozhe Xu$^{*1,4}$, Cheng Cheng$^{*2}$, Hongrui Sang$^{\dagger3}$, Zhipeng Wang$^{1,4}$\\
Qiyong He$^{1,2}$, Xiuxian Li$^{1,2}$, and Bin He$^{1,4}$
	\thanks{$^{*}$ Equal Contributions}
    \thanks{ $^{\dagger}$ corresponding author(e-mail: hrsang@shmtu.edu.cn)}
    \thanks{This work was supported in part by the National Natural Science Foundation of China under Grant 62088101, Grant 62403366 and Grant 62473294; in part by Natural Science Foundation of Shanghai under Grant 25ZR1402185; in part by Shanghai Municipal Science and Technology Major Project under Grant 2021SHZDZX0100; in part by the University-Industry-Research Innovation Fund of China under grant 2024ZY026.}
	\thanks{$^{1}$College of Electronics and Information Engineering, Tongji University, Shanghai 201804, China. $^{2}$Frontiers Science Center for Intelligent Autonomous Systems, Shanghai 201210, China. $^{3}$School of Logistics Engineering, Shanghai Maritime University, Shanghai 201306, China. $^{4}$State Key Laboratory of Autonomous Intelligent Unmanned Systems, Shanghai 201210, China.
}
}

\begin{document}

	\maketitle
	\thispagestyle{empty}
	\pagestyle{empty}

\begin{abstract}		
Autonomous docking between Unmanned Aerial Vehicles (UAVs) and ground robots is essential for heterogeneous systems, yet most existing approaches target wheeled platforms whose limited mobility constrains exploration in complex terrains. Quadruped robots offer superior adaptability but undergo frequent posture variations, making it difficult to provide a stable landing surface for UAVs. To address these challenges, we propose an autonomous UAV–quadruped docking framework for GPS-denied environments. On the quadruped side, a Hybrid Internal Model with Horizontal Alignment (HIM-HA), learned via deep reinforcement learning, actively stabilizes the torso to provide a level platform. On the UAV side, a three-phase strategy is adopted, consisting of long-range acquisition with a median-filtered YOLOv8 detector, close-range tracking with a constraint-aware controller that integrates a Nonsingular Fast Terminal Sliding Mode Controller (NFTSMC) and a logarithmic Barrier Function (BF) to guarantee finite-time error convergence under field-of-view (FOV) constraints, and terminal descent guided by a Safety Period (SP) mechanism that jointly verifies tracking accuracy and platform stability. The proposed framework is validated in both simulation and real-world scenarios, successfully achieving docking on outdoor staircases higher than 17 cm and rough slopes steeper than 30 degrees. Supplementary materials and videos are available at: \url{https://uav-quadruped-docking.github.io}.
\end{abstract}

	\section{INTRODUCTION}

Heterogeneous cooperative systems integrating Unmanned Aerial Vehicles (UAVs) and Unmanned Ground Vehicles (UGVs) expand operational scope and improve efficiency compared with single-domain platforms \cite{8794265}. Autonomous docking is a key capability in such collaboration, yet most existing approaches focus on wheeled UGVs whose mobility is limited to flat terrain, restricting deployment in complex environments. Moreover, dynamic docking requires UAVs to achieve precise localization and safe landing on moving platforms, placing strict requirements on sensor fusion and robust control \cite{CHENG20222788}.

\begin{figure}[t]
	\centering
	\includegraphics[width=0.8\linewidth]{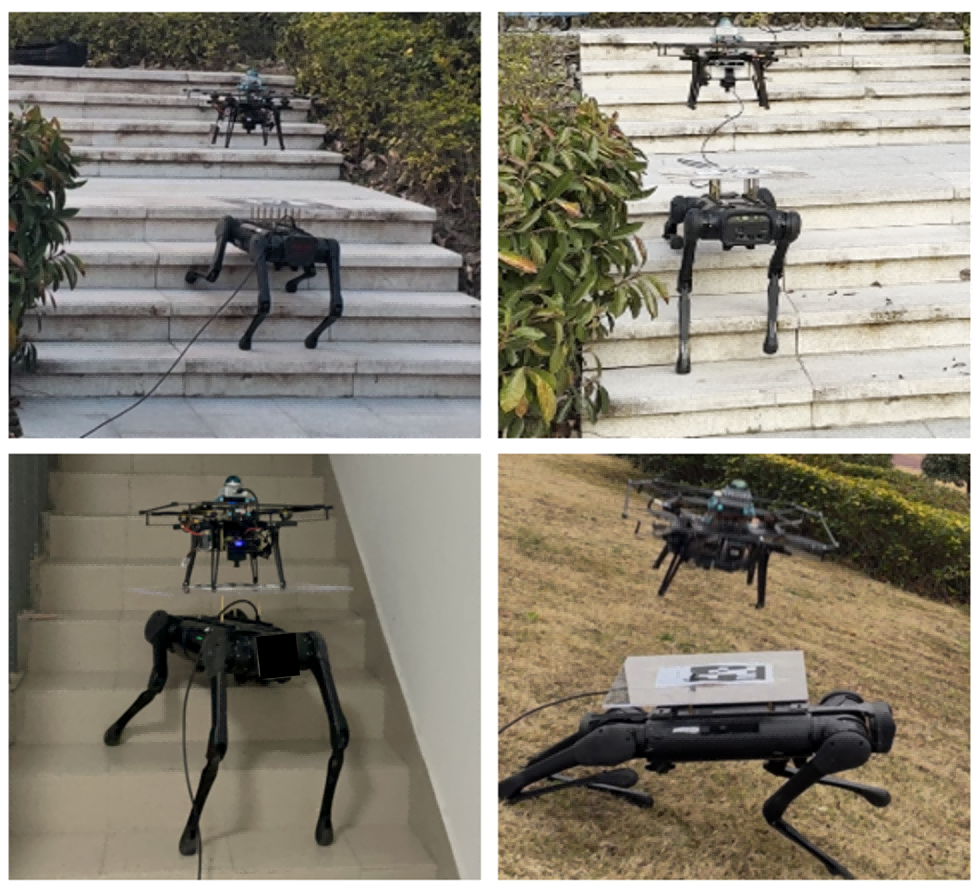} 
	\caption{Docking experiments in complex terrains. The top-left and top-right show tests on a $17\,\text{cm}$ outdoor stair where the quadruped actively adjusts its posture while ascending and descending. The bottom-left shows an indoor $15\,\text{cm}$ stair experiment, and the bottom-right presents docking on a $30^{\circ}$ outdoor rough slope, demonstrating the robustness of the proposed method.}
	\label{fig:intro} 
\end{figure}

Quadruped robots outperform wheeled and tracked UGVs in unstructured terrains and enable UAV collaboration in challenging environments such as mountains or tunnels. Recent advances in deep reinforcement learning (DRL) provide a new paradigm for quadruped locomotion \cite{hwangbo2019learning,lee2020learning,hoeller2024anymal}. Compared with model-based controllers (e.g., Model Predictive Control, MPC \cite{di2018dynamic}), DRL maps observations directly to joint actions, enabling strong adaptability to unseen terrains and sensor noise. However, prior work mainly focuses on single-robot locomotion and rarely considers collaborative tasks that impose posture and velocity constraints for UAV docking.

This work studies UAV–quadruped docking in complex three-dimensional terrains where large posture variations occur. On stairs higher than 15 cm or slopes steeper than $20^{\circ}$, quadruped pitch angles can exceed $20^{\circ}$, producing unstable landing surfaces for UAVs. Such instability also degrades vision-based tracking (e.g., AprilTag), making it necessary for UAVs to fuse onboard sensing with quadruped state feedback for robust tracking and accurate localization.

\subsection{Related Works}

\textbf{Autonomous docking}  
Autonomous docking techniques can be broadly categorized into infrastructure-based and on-board sensor–based methods. Infrastructure-assisted schemes typically rely on external positioning systems such as GPS, BeiDou, or motion capture systems (MCS) \cite{9044775,XIA2022105288}. For example, LiDAR, IMU, and GPS were tightly fused in \cite{10726742} to improve environment perception and landing point selection for powered parachute UAVs, while \cite{Priambodo_2022} combined GPS and ArUco marker detection for landing in simulation. However, these methods depend heavily on infrastructure, incur high deployment costs, and perform poorly when GPS signals are degraded or unavailable in cluttered environments such as vegetation, water bodies, or indoor spaces \cite{balamurugan2016survey}.

On-board sensor–based docking methods \cite{8793851,10436548} employ vision, LiDAR, and IMU to operate in GPS-denied environments. In \cite{9623487}, a lightweight vision-based controller enabled quadrotors to land on unknown moving platforms. In \cite{10472074}, semantic and depth information from cameras and LiDAR were fused to autonomously search for safe landing sites. Nevertheless, these approaches remain sensitive to sensor quality, degrade in low-light or adverse conditions, and often introduce significant computational overhead.

Most existing docking approaches target wheeled UGVs, whose mobility is limited to flat or structured terrain. In rugged environments such as rubble fields, steep slopes, or stairs, wheeled platforms struggle to operate effectively, and posture variations of the landing surface further increase the risk of field-of-view (FOV) loss during docking.

\textbf{Learning-based quadruped locomotion}  
Quadruped robots outperform wheeled UGVs in unstructured terrains and are increasingly adopted for exploration tasks. Reinforcement learning (RL) has emerged as an effective paradigm for quadruped locomotion \cite{hwangbo2019learning,hoeller2024anymal,rudin2022learning}, with policies trained in simulation and transferred to hardware via zero-shot sim-to-real. In \cite{kumar2021rma}, an adaptation module within a teacher–student framework enabled robust locomotion, while \cite{margolis2023walk} incorporated behavior parameters (e.g., body height, body pitch, and foot swing height) into RL observations and rewards to produce versatile locomotion behaviors. However, these approaches mainly address mildly uneven terrains and lack active posture control in complex environments.

Recent RL-based methods demonstrate strong adaptability in unstructured 3D terrains. For instance, \cite{nahrendra2023dreamwaq} introduced a context-aided estimator to jointly infer body state and environmental context, while \cite{long2024hybrid} combined internal model control (IMC) \cite{rivera1986internal} with reinforcement learning to propose a Hybrid Internal Model (HIM) for velocity estimation and system response modeling, enabling agile locomotion over stairs and slopes.

Nevertheless, most RL frameworks treat body posture alignment as a passive outcome of terrain adaptation. In UAV–quadruped collaborative tasks such as autonomous docking, quadrupeds must actively regulate posture to provide a level landing surface for UAVs. Existing methods lack such task-oriented posture control, limiting their applicability in aerial–ground cooperation.

	\begin{figure*}[t]
		\centering
        \vspace{2mm}
		\includegraphics[width=0.8\linewidth]{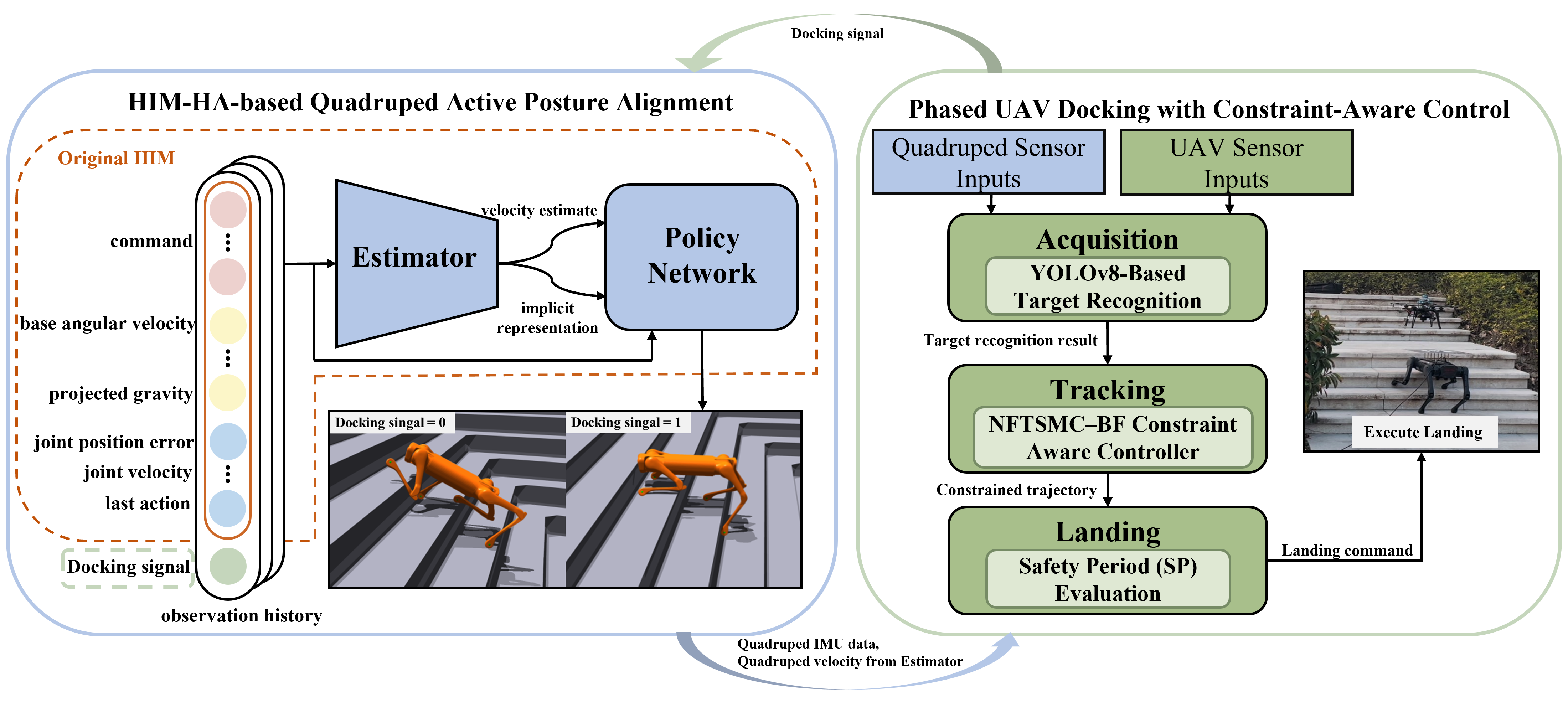} 
		\caption{An overview of our UAV-Quadruped collaborative autonomous docking scheme. Both the UAV and the quadruped robot actively participate in the docking process, transferring data information between them. The UAV sends the docking request “docking signal” to the quadruped robot, and the quadruped robot actively aligns the posture after receiving the signal, and sends the IMU readings together with velocity estimation derived from the Estimator to the UAV.}
		\label{fig:overview} 
	\end{figure*}

	\subsection{Contributions}	
	To address the landing difficulties caused by the dynamic torso posture of the quadruped robot, we model the UAV-quadruped robot docking task in complex terrain environments as a heterogeneous robot cooperative planning problem. Optimization methods are designed separately for the UAV and the quadruped robot to achieve safe and precise docking.
	Our contributions are as follows:

		\begin{itemize}
	
		\item We propose an autonomous UAV–quadruped docking framework that integrates multi-sensor fusion with an active posture alignment mechanism to enhance reliability in GPS-denied environments.
		\item On the quadruped side, we extend the Hybrid Internal Model (HIM) and introduce HIM with Horizontal Alignment (HIM-HA), a reinforcement learning–based locomotion strategy that enables quadrupeds to actively stabilize their torso and provide a level landing platform in complex 3D terrains.
        \item On the UAV side, we develop a constraint-aware controller that integrates a Nonsingular Fast Terminal Sliding Mode Controller (NFTSMC) with a logarithmic Barrier Function (BF) to guarantee finite-time error convergence under field-of-view (FOV) constraints, and introduce a Safety Period (SP) strategy to ensure robust and reliable terminal descent.
		\item We validate the proposed framework in both simulation and real-world experiments across challenging environments, including stairs and steep slopes. To the best of our knowledge, this is the first study to demonstrate successful UAV–quadruped docking in unstructured terrains.
		
	\end{itemize}
	
	\section{METHOD}

	\subsection{Overview}

An overview is presented in Fig. \ref{fig:overview}. The docking scheme consists of two parts: the HIM-HA-based active posture alignment strategy for a quadruped robot and the phased UAV docking with constraint-aware NFTSMC-BF control.

On the quadruped robot side, we design a HIM-HA method based on HIM that actively aligns the torso horizontally in complex terrain. HIM-HA maintains the robust traversal capability of HIM, while sharing the same network structure with differences only in observation inputs and the training process. During docking, after receiving the docking request signal from the UAV, the quadruped aligns its torso horizontally according to the policy output to provide a safe landing platform.

On the UAV side, the docking process is organized into three phases. In the acquisition phase, a YOLOv8-based detector recognizes the quadruped beyond the range of fiducial markers. In the tracking phase, a constraint-aware controller integrates NFTSMC-BF to ensure finite-time error convergence under FOV constraints. In the landing phase, an SP strategy jointly evaluates UAV tracking performance and quadruped stability to guarantee safe and reliable descent.

\subsection{HIM-HA-based Quadruped Active Posture Alignment}
	
	This paper addresses the problem of aerial-ground docking in complex terrestrial environments, requiring a quadruped  robot to explore challenging three-dimensional terrains freely. The HIM method \cite{long2024hybrid} demonstrates relatively robust motion performance across various complex environments. The original HIM method is based on the principle of IMC \cite{rivera1986internal}, treating external states as disturbances that can be estimated from the system response of the robot. Consequently, the HIM method exhibits robustness to terrain variations and external disturbances. The core of the HIM method is to integrate a contrastive learning-based estimator into the reinforcement learning paradigm to estimate the robot's velocity and system response representation.
	
	In our collaborative docking scheme, the UAV requires the quadruped robot to provide a dynamically stable horizontal reference plane for landing. However, conventional HIM algorithms focus on motion stability, where the torso posture is passively generated through a terrain-adaptive mechanism, lacking an active horizontal alignment capability. To enable the quadruped robot to provide suitable landing conditions for the UAV on complex terrains such as stairs and rough slopes, we extend HIM by incorporating the Horizon Alignment functionality, resulting in the HIM-HA method. This enhancement allows the quadruped  robot to actively align its posture horizontally in complex scenarios.
	
	The HIM-HA method retains the same network architecture as HIM. Like HIM, we model the legged locomotion task as a sequential decision problem. The observation $o_{t}$ of the HIM module includes the desired velocity, proprioceptive information from its joint encoder and IMU, as well as the last action. As illustrated in Fig. \ref{fig:overview}, the only modification made to the observation of HIM-HA $o_{t}^{HA} $ compared to $o_{t}$ is the addition of a one-dimensional docking signal value $\textit{D}$ at the end of each observation vector. This signal takes binary values of either $0$ or $1$,  which simulates the signal sent by the UAV when entering the docking mode during real-world experiments. Specifically, $\textit{D}=1$ indicates that the quadruped enters the docking mode, during which it decreases locomotion speed and actively maintains a level posture, while $\textit{D}=0$ denotes standard locomotion with no additional posture regulation. The HIM/HIM-HA framework processes observation information from six consecutive time steps and inputs it into the framework to output the bias between the target joint position and nominal joint position. Then, the torques for all joints are obtained through the PD controller. Additionally, the estimator module within the HIM/HIM-HA framework provides velocity estimation of the quadruped robot. During docking tasks, this estimated velocity is transmitted to the UAV to assist in predicting the quadruped robot's future positions.
	
	Compared to the original HIM, we introduce an additional reward and propose a progressive curriculum learning method tailored to the task in this paper.
	
	\textbf{Task-conditioned Horizontal Alignment Reward Design} In the original reward formulation, a general orientation penalty was introduced by minimizing the squared horizontal projection of the gravity vector expressed in the robot body frame:
\begin{equation}
r_{\text{ori}} = g_x^2 + g_y^2 ,
\end{equation}
where $(g_x,g_y,g_z)$ denotes the normalized gravity direction in the body frame. This term discourages large roll and pitch deviations, but is weighted with a small coefficient to avoid over-constraining the robot on challenge terrains.
For the UAV docking task, however, we require the quadruped to actively maintain a horizontally aligned base. We therefore design a task-conditioned \emph{Horizontal Alignment (HA) reward}, which is only strongly activated when the docking mode indicator $\textit{D} \in \{0,1\}$ is enabled:
\begin{equation}
    r_{\text{HA}} = \alpha(\textit{D})\,\big(\max(g_x^2+g_y^2 - \delta,\,0)\big) 
                  + \beta(\textit{D})\,(\omega_x^2+\omega_y^2),
\end{equation}
where $\omega_x,\omega_y$ denote roll and pitch angular velocities, and $\delta$ is a small deadzone threshold below which minor deviations are ignored.
The coefficients $\alpha(\textit{D})$ and $\beta(\textit{D})$ are defined in a piecewise manner:
\begin{equation}
\alpha(\textit{D}),\, \beta(\textit{D}) =
\begin{cases}
\alpha_0,\, \beta_0, & \textit{D}=0 \\
\alpha_1,\, \beta_1, & \textit{D}=1
\end{cases}
\end{equation}

Here, $\alpha_0$ and $\beta_0$ are set to small values so that the HA reward degenerates into a weak orientation prior during normal locomotion, thus avoiding overly conservative behavior on sloped terrains. In contrast, $\alpha_1$ and $\beta_1$ are assigned significantly larger values when $\textit{D}=1$, forcing the quadruped to prioritize roll/pitch stabilization and suppression of oscillatory motion to provide a reliable horizontal platform for UAV landing. The value of $\textit{D}=1$ is randomly sampled during training.

	\textbf{Curriculum Learning Design}  Forcing a robot to move in a specific posture across any terrain is often more challenging to train than without such constraints. To facilitate stable convergence and robust generalization, we employ a two-fold curriculum learning strategy, consisting of curriculum commands and curriculum terrains. Importantly, both curricula are conditioned on the docking mode indicator $\textit{D} \in \{0,1\}$, which specifies whether the quadruped is required to act as a horizontally aligned platform for UAV landing.

For the command curriculum, the velocity commands are initially sampled from a narrow range, such that the robot only needs to perform simple low-speed forward walking behaviors. As training progresses, in the case of $\textit{D}=0$, the command ranges are gradually expanded to cover higher linear and angular velocities, enabling the policy to acquire diverse and agile locomotion skills. In contrast, when $\textit{D}=1$, the command ranges remain restricted to small magnitudes (e.g., limited forward speed and minimal yaw rotation), ensuring that the policy focuses on maintaining a horizontal torso rather than pursuing agility.

For the terrain curriculum, the progression strategy is differentiated as well. In the case of $\textit{D}=0$, the robot is first trained on flat ground and then gradually exposed to more challenging terrains, including slopes, stairs, and uneven surfaces, in order to develop general locomotion robustness. In contrast, when $\textit{D}=1$, the terrain progression is deliberately slowed down: the robot initially practices maintaining horizontal alignment on relatively simple terrains, and only after achieving sufficient stability does it encounter steeper slopes and higher stairs. Moreover, in docking mode, the Horizontal Alignment (HA) reward plays a dominant role, ensuring that the policy prioritizes torso stabilization over velocity tracking while adapting to progressively harder terrains.

\subsection{ Phased Autonomous Docking Mission Planning}
\begin{figure}
    \centering
    \vspace{2mm}
    \includegraphics[width=0.9\linewidth]{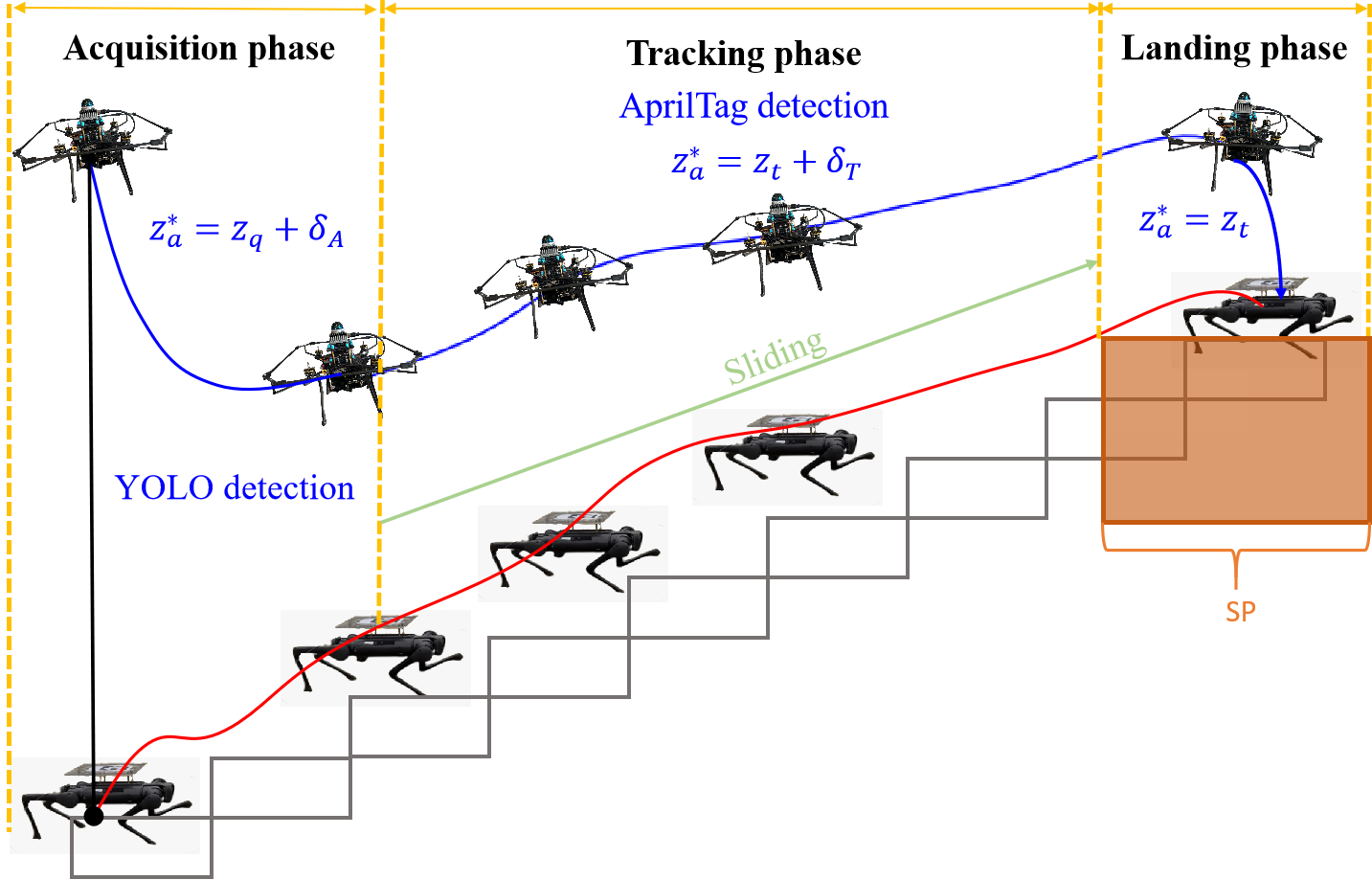}
    \caption{ Illustration of the three docking phases: acquisition, tracking, and landing.}
    \label{fig:S3}
\end{figure}

To achieve safe and reliable tracking and landing on a dynamic ground target, we decompose the entire autonomous mission flow into three sequential phases  as shown in Fig.~{\ref{fig:S3}}: the acquisition phase (long-range identification and tracking), the tracking phase (close-range precision following), and the landing phase (steady-state discrimination and landing).

 {\bf Acquisition phase} The initial acquisition phase of autonomous docking missions is dedicated to the robust detection and maneuvering of a UAV toward a target from a significant stand-off distance. While fiducial markers such as AprilTags provide high-precision pose estimation for close-range operations, their performance degrades significantly at extended distances. Signal degradation due to low pixel resolution, partial occlusions, and intermittent line-of-sight—compounded by scenarios where the target, such as a quadruped robot, is not fully visible—renders marker-based approaches unreliable for the critical initial acquisition phase. This unreliability necessitates a perception strategy that does not depend on high-resolution, consistently visible markers for initial engagement.

To address these limitations, this paper presents a novel tracking methodology that integrates a deep learning-based detector with a robust filtering scheme for stable state estimation. Specifically, a YOLOv8 convolutional neural network, fine-tuned for quadruped robot detection, serves as the primary perception module. The network processes the onboard video stream in real-time to identify the quadruped within the UAV's FOV. A detection is considered valid only if its associated confidence score exceeds a predefined threshold, $\tau_{conf}$, effectively filtering out ambiguous, low-quality detections common in long-range imagery. A key contribution of our approach is the introduction of a median filter applied to the temporal sequence of the quadruped's 2D image-plane coordinates. This filtering strategy effectively rejects sporadic, high-variance outliers inherent to long-range detection, providing a stabilized estimate of the target's centroid, $\mathbf{p}_{img} = [u, v]^T$. 

The UAV’s current position $\mathbf{p}_a = [x_a, y_a, z_a]^T \in \mathbb{R}^3$ is measured in the world frame via onboard sensors. The stabilized 3D quadruped position $\mathbf{p}_q$ estimate  serves as the dynamic setpoint for the UAV's guidance system. A Proportional-Integral (PI) position controller is implemented to generate velocity commands that minimize the error  current position. The commanded velocity $\mathbf{v}_{cmd}$, is calculated as:
\begin{equation}\mathbf{v}_{cmd} = K_p (\mathbf{p}_q - \mathbf{p}_{a}) + K_i \int (\mathbf{p}_q - \mathbf{p}_{a}) dt,\end{equation}
where $K_{p}$ and $K_{i}$ are the proportional and integral gain matrices, respectively. The integral term is crucial for eliminating steady-state errors and ensuring precise convergence. These commands are transmitted directly to a Pixhawk 4 flight control unit, enabling real-time, closed-loop trajectory tracking. This strategy focuses on efficiently approaching the quadruped, and at this moment, the UAV height is $z_a^* = z_q + \delta_A, \; \delta_A > 0$.

{\bf{Tracking phase}} Upon successful completion of the long-range acquisition phase, the UAV transitions to the close-range tracking phase. The core task of this phase is to precisely track the AprilTag marker located on the quadruped robot's deck in preparation for the final autonomous landing. In contrast to the acquisition phase, this stage imposes more stringent requirements on tracking precision and constraint satisfaction. The target marker position, $\mathbf{p}_{t} = [x_t, y_t, z_t]^T$, is estimated by the onboard camera using the AprilTag detection system. The tracking error vector is defined as 
\begin{equation}\mathbf{e} = \mathbf{p}_{a} - \mathbf{p}_{t} = [e_x, e_y, e_z]^T.
\end{equation}

This primary objective must be achieved while strictly satisfying the following state constraints for all time $t \ge 0$:

\begin{enumerate}
    \item FOV: To ensure continuous target pose estimation, the horizontal relative displacement between the UAV and the target must be confined within a predefined safe radius $d_{s}$.
    \begin{equation} e_x^2(t) + e_y^2(t)\le d_{s}^2. \end{equation}

    \item Collision Avoidance: To guarantee operational safety, the UAV's altitude $z_{a}(t)$ must be maintained above a minimum safe altitude $z_{min}$ to avoid collision with the target deck or the ground.
    \begin{equation} z_{a}^{*}(t) =  z_{q}(t)+\delta_T.\end{equation}
\end{enumerate}

To achieve finite-time convergence of the tracking error and strictly enforce operational constraints, this section presents a NFTSMC-BF. This approach overcomes the asymptotic convergence limitation of linear sliding mode controllers, which is critical for the high-precision terminal phase of the tracking mission.

We design the definition of an sliding surface:
\begin{equation}\mathbf{s} = \dot{\mathbf{e}} + \alpha \mathbf{e} + \beta \mathbf{e}^{p/q}, \end{equation}
 where $\mathbf{s} \in \mathbb{R}^3$ is the sliding variable; $\alpha, \beta \in \mathbb{R}^{3\times3}$ are positive definite diagonal gain matrices; and $p, q$ are odd positive integers satisfying $1 < p/q < 2$. 

To strictly enforce the FOV constraint, a logarithmic BF $B(\mathbf{e})$ is incorporated, defined as:
\begin{equation}B(\mathbf{e}) = \log\left(\frac{1}{d_{s}^2 - (e_x^2 + e_y^2)}\right),\end{equation}
where $d_{s}$ is a conservative radius that defines the safe tracking region. The gradient of this function, $\nabla B(\mathbf{e})$, generates a repulsive potential that prevents the state from reaching the constraint boundary.

Based on sliding dynamics and the UAV model, the control law $\mathbf{u}$  is synthesized to ensure the properties of reach, sliding, and convergence. The resulting controller is given by:
\begin{equation}\begin{aligned} \mathbf{u} =& {\mathbf{M}}\left( \ddot{\mathbf{p}}_{t} - \alpha\dot{\mathbf{e}} - \beta\frac{p}{q}\text{diag}\left(|\mathbf{e}|^{\frac{p}{q}-1}\right)\dot{\mathbf{e}} \right)  + {\mathbf{G}}  \\& - K_{d}\mathbf{s} - K_{sw}\text{sgn}(\mathbf{s})- K_{b}\nabla B(\mathbf{e}),\end{aligned}\end{equation}
where $ {\mathbf{M}}$ and $ {\mathbf{G}}$ are the   UAV's inertia matrix and gravitational/aerodynamic forces, respectively.   The controller is composed of: an equivalent control component that cancels the system dynamics and compensates for the target's motion and the nonlinear sliding surface dynamics; a robustifying component with a proportional reaching term $-K_{d}\mathbf{s}$ and a signum function switching term $-K_{sw}\text{sgn}(\mathbf{s})$ to ensure finite-time convergence to the sliding surface despite uncertainties; and a repulsive term $-K_{b}\nabla B(\mathbf{e})$ to enforce the state constraints.

{\bf{Landing phase}} 
To ensure a robust landing decision, the system requires that both the UAV’s tracking performance and the stability of the target platform satisfy predefined criteria. Inspired by \cite{huang2021linear}, we define that the system enters a SP and initiates the landing procedure (i.e., updating the altitude setpoint to $z_{a}^* = z_t$) if and only if the following two independent conditions are simultaneously satisfied within the recent time window $[t - T_s, t]$.
\begin{equation}
    \sqrt{\frac{1}{T_s}\int_{t-T_s}^{t} \|\mathbf{e}(\tau)\|^2 d\tau} \le \epsilon_p,
\end{equation}
\begin{equation}
    \frac{1}{T_s}\int_{t-T_s}^{t} \left( w_{\omega}\|\boldsymbol{\omega}_t(\tau)\| + w_{\phi}|\phi_t(\tau)| + w_{\theta}|\theta_t(\tau)| \right) d\tau \le \epsilon_t,
\end{equation}
where $\boldsymbol{\omega}_t$ is the target angular velocity, $\phi_t$ and $\theta_t$ are the target roll and pitch angles, respectively. The constants $\omega_{\omega}$, $\omega_{\phi}$, and $\omega_{\theta}$ are the corresponding weighting coefficients that reflect the relative importance of angular velocity, roll, and pitch deviations in the landing initiation evaluation.  $\epsilon_p$ and $\epsilon_t$ are the upper bound constants satisfied under SP, respectively. This initiation mechanism employs a dual-verification logic to comprehensively evaluate the prerequisites for landing. 

		\section{EXPERIMENT RESULTS}
        \subsection{Experimental Platforms}
\begin{figure}[t]
	\centering
    \vspace{2mm}
	\includegraphics[width=0.8\linewidth]{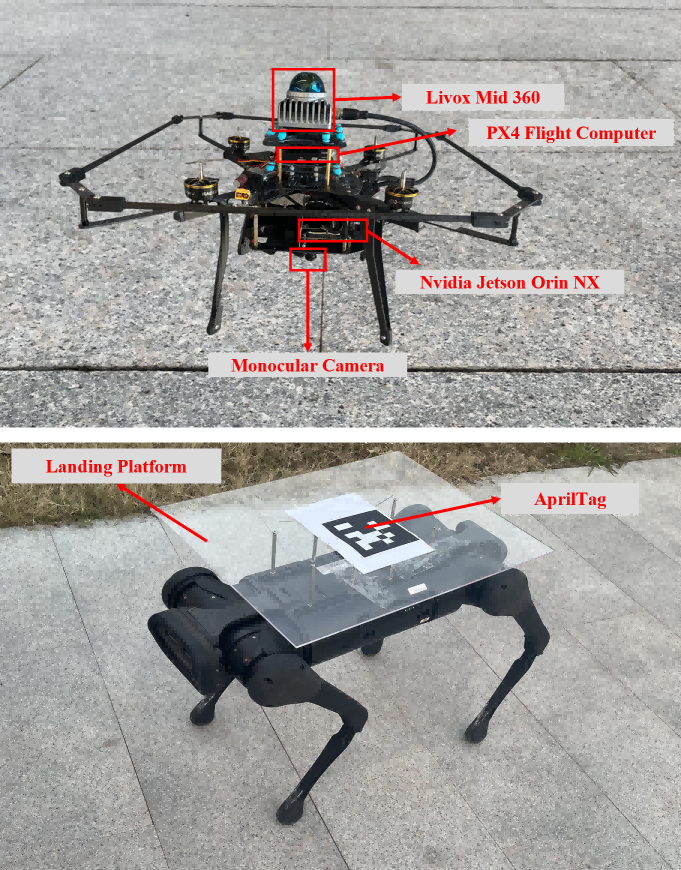} 
	\caption{Our experimental platform for UAV–quadruped docking, showing the quadrotor UAV equipped with onboard sensors and the Unitree quadruped robot with an AprilTag landing plate.}
	\label{fig:platform} 
\end{figure}

As shown in Fig. \ref{fig:platform}, the UAV platform employs a self-designed quadrotor architecture with a main body dimension of 	$44 \times 44 \,\ \text{cm}^2$,  equipped with a landing gear system (deployment area $24 \times 24 \,\ \text{cm}^2$) to ensure ground stability.   The perception system consists of a Livox Mid-360 LiDAR module mounted on the top for high-precision 3D environmental perception, and a monocular vision sensor integrated at the bottom  dedicated to AprilTag landing marker recognition.   The flight control system incorporates a PX4 open-source autopilot system, which utilizes a built-in 6-axis MEMS IMU  for precise attitude estimation.   The edge computing unit features an NVIDIA Jetson Orin NX embedded platform, providing computational support for real-time object recognition, sensor fusion, and autonomous docking algorithms.

We utilize the Unitree Aliengo quadruped robot as our ground robotic platform. The robot's body dimensions are $65 \,\ \text{cm}$ in length and $31 \,\ \text{cm}$ in width, with a standing height of $60 \,\ \text{cm}$. A $50\times40 \,\ \text{cm}^2$ acrylic plate is mounted on the upper part of its torso, serving as the landing platform. An AprilTag is positioned at the center of the platform for visual recognition. We deploy and run the HIM-HA policy on a laptop computer and control the quadruped robot motion over a network cable. During the docking phase, the quadruped robot and the UAV communicate via WiFi.

\begin{table*}[htbp]
	\centering
    \vspace{2mm}
	\caption{Baseline and ablation of quadruped posture alignment in Simulation}
    \renewcommand{\arraystretch}{1.5} 
	\begin{tabular}{ccccccccc}
		\cline{1-7}
		\multicolumn{1}{l|}{}                    & \multicolumn{2}{c|}{Average Terminated Difficulty Level} & \multicolumn{2}{c|}{Average Displacement(m)}       & \multicolumn{2}{c}{Success Rate}            &  &  \\[0.3em]
		\multicolumn{1}{l|}{}                    & \qquad  Stair                      & \multicolumn{1}{c|}{Slope}  & Stair                & \multicolumn{1}{c|}{Slope}  & Stair                & Slope                &  &  \\ \cline{1-7}
		\multicolumn{1}{c|}{RMA}                 &\qquad{5.3}   & \multicolumn{1}{c|}{4.2}    & 40.180               & \multicolumn{1}{c|}{33.552} & 0.49                    & 0.40                    &  &  \\
		\multicolumn{1}{c|}{WTW}                 &\qquad{5.1}   & \multicolumn{1}{c|}{4.2}    & 36.443               & \multicolumn{1}{c|}{31.044} & 0.46                    & 0.38                    &  &  \\
		\multicolumn{1}{c|}{WBC}                 &\qquad{8.4}   & \multicolumn{1}{c|}{7.7}    & 65.554               & \multicolumn{1}{c|}{59.338} & 0.82                    & 0.78                    &  &  \\
		\multicolumn{1}{c|}{Ours w/o $r_{\text{HA}}$} &\qquad{7.4}   & \multicolumn{1}{c|}{6.1}    & 57.784               & \multicolumn{1}{c|}{44.271} & 0.71                    & 0.57                    &  &  \\
		\multicolumn{1}{c|}{Ours w/o Curriculum} &\qquad{5.9}   & \multicolumn{1}{c|}{4.6}    & 49.665               & \multicolumn{1}{c|}{31.935} & 0.60                    & 0.41                    &  &  \\
		\multicolumn{1}{c|}{Ours}                & \qquad{8.8}   & \multicolumn{1}{c|}{7.9}    & 71.602               & \multicolumn{1}{c|}{67.339} & 0.87                  & 0.79                  &  &  \\ \cline{1-7}
		& \multicolumn{1}{l}{}       & \multicolumn{1}{l}{}        & \multicolumn{1}{l}{} & \multicolumn{1}{l}{}        & \multicolumn{1}{l}{} & \multicolumn{1}{l}{} &  &  \\
		& \multicolumn{1}{l}{}       & \multicolumn{1}{l}{}        & \multicolumn{1}{l}{} & \multicolumn{1}{l}{}        & \multicolumn{1}{l}{} & \multicolumn{1}{l}{} &  &  \\
			\end{tabular}\label{ablation study}
			\end{table*}

        \subsection{Quadruped Posture Alignment in Simulation}
The training of the HIM-HA algorithm is conducted in Isaac Gym. The network architecture can be referenced in \cite{long2024hybrid}. When implementing our approach, it is crucial to employ curriculum learning to reduce exploration difficulty while maintaining effective locomotion over complex terrains and simultaneously aligning the robot's body posture.

\begin{figure}[t]
			\centering
            \vspace{2mm}
			\includegraphics[width=0.98\linewidth]{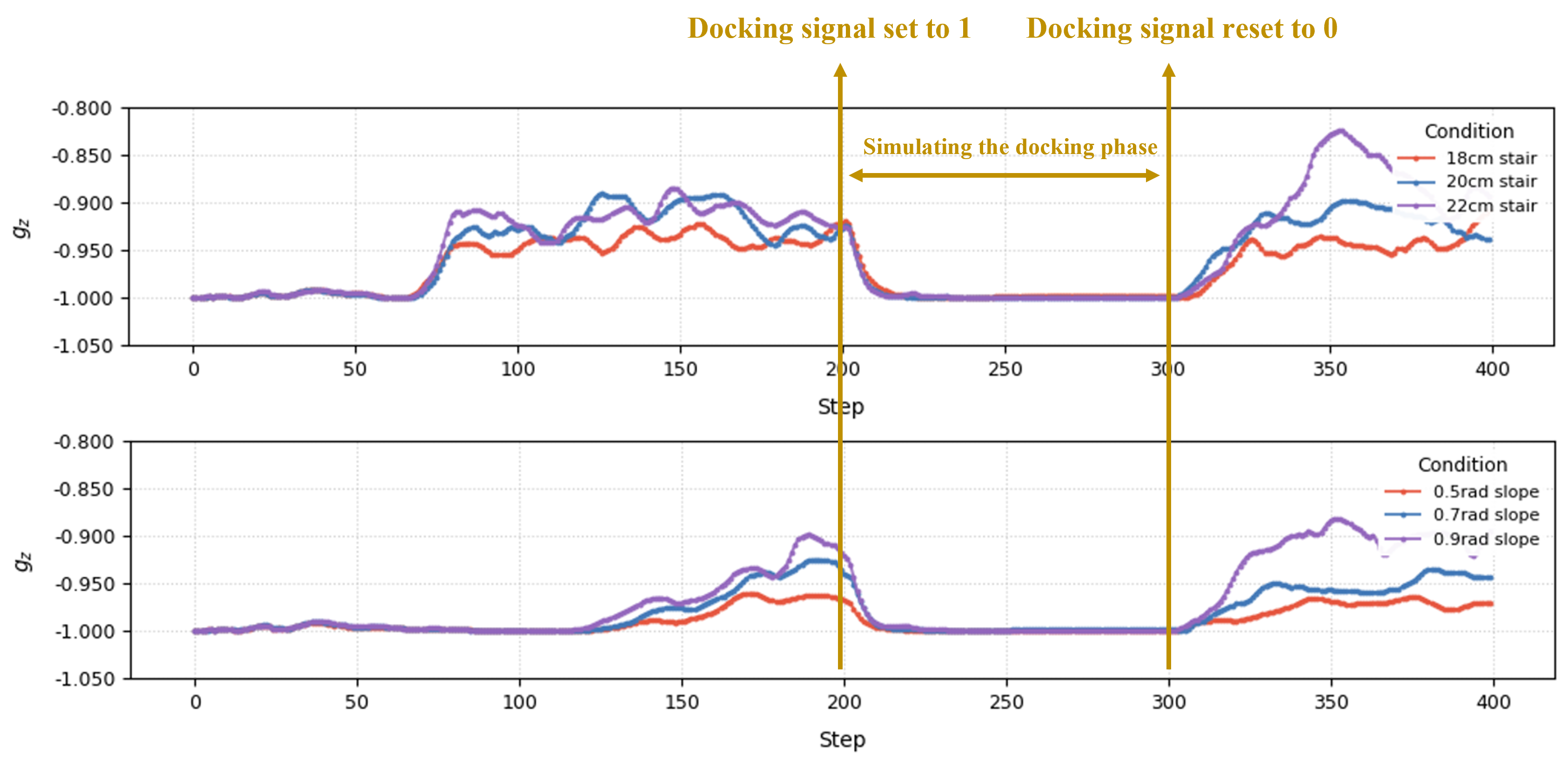} 
			\caption{HIM-HA Policy Testing in Different Terrains. We conducted tests on the HIM-HA method's capability to adjust the robot’s posture under various terrain conditions with different difficulty levels in simulation (Top: stair, Bottom: slope). At step 200, the robot receives a docking signal, requiring it to adjust its posture to a horizontal state. At step 300, the robot is instructed to restore its original posture. The vertical axis $g_{z}$ represents the projection of the normalized gravity vector onto the body-frame $z$-axis. Values closer to $-1$ indicate that the robot’s torso is more horizontally aligned with the ground.}
			\label{fig:simulation_1} 
		\end{figure}

    	\textbf{Testing in Different Terrains}	Fig.	\ref{fig:simulation_1} demonstrates the capability of the proposed method to actively adjust the robot's posture to maintain a horizontal orientation while traversing diverse staircases and sloped terrains. The robot starts from flat ground and then moves onto ascending steps or slopes. The vertical axis $g_{z}$ represents the projection of the gravity vector onto the z-axis of the robot's base coordinate system. At the 200th time step, a docking signal is issued to the robot, setting the observation $\textit{D}$ corresponding to HIM-HA to 1. Regardless of the type or complexity of the terrain, the quadrupedal robot can rapidly adjust its torso posture to maintain a horizontal orientation (evidenced by gravity projection $g_{z}$ approaching -1). Results demonstrate that the HIM-HA method can provide landing conditions for UAVs even on $22\,\ \text{cm}$ stairs and slopes exceeding $50^\circ$ in simulation.

        \textbf{Baseline and Ablation} To verify the efficiency of the HIM-HA method, we constructed a continuous track to compare different approaches. There are two types of tracks. The first type comprises staggered ascending and descending steps, while the second type consists of staggered ascending and descending slopes. The terrain difficulty increases progressively along the track. The initial step height is set at 10 cm, with each difficulty level increasing by 1 cm. For the slope track, the initial incline is 0.15 rad, with each difficulty level increasing by 0.05 rad. Each terrain section spans 8 meters (e.g., an ascending and descending slope of 4 meters constitutes a single difficulty level in the slope track). There are ten consecutive sections of such alternating slope terrain, meaning that each track consists of ten levels of difficulty, resulting in a total track length of 80 meters.

        We compare our parkour policy with several baselines and ablation studies. The compared settings are as follows:
		\begin{itemize}
		\item \textbf{RMA}: We train an RMA-based RL policy following the methodology in \cite{kumar2021rma} and enhance it by incorporating a Horizontal Alignment Reward coupled with a simplified terrain curriculum.

        \item \textbf{WTW}: We train a ''Walk These Ways" RL policy following \cite{margolis2023walk} and modify its pitch tracking command and the corresponding reward terms to respond to the docking signal.

        \item \textbf{WBC}: We implement a Whole-Body Control (WBC) baseline that stabilizes the quadruped’s torso by solving a quadratic program (QP) at each timestep in isaacgym.

        \item \textbf{w/o task-conditioned horizontal alignment reward $r_{\text{HA}}$}: We replace the horizontal alignment reward $r_{\text{HA}}$ with the orientation penalty $r_{\text{ori}}$, applied with a larger weight during training.
        \item \textbf{w/o curriculum}: Training without curriculum learning.

        \item \textbf{Our method}.

	\end{itemize}

    All methods operate under a forward velocity command of 0.8 m/s. Every 5 seconds, a docking signal is received, at which point the forward velocity of all robots is set to 0.3 m/s. The parameter $\textit{D}$ is reset to 0 three seconds later. Each strategy is tested on the track 200 times, and we record the following metrics: the maximum terrain difficulty level reached, the distance traveled before the robot falls, and whether the robot successfully reaches 70\% of the total track length.

        \begin{figure}[t]
			\centering
			\includegraphics[width=0.8\linewidth]{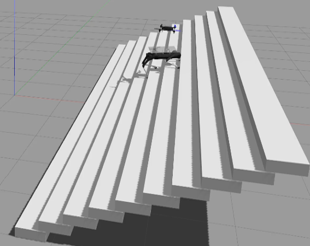} 
			\caption{The Gazebo simulation environment.}
			\label{fig:sim_gazebo} 
		\end{figure}

    Table \ref{ablation study} presents the performance of different methods on the tracks. The results demonstrate that our proposed method outperforms all other approaches. Our method exhibits degraded performance in extreme terrains when either the Horizontal Alignment Reward is removed or curriculum learning is abandoned. The unsatisfactory results of RMA and WTW stem from their limited capability to identify latent terrain features compared to HIM. We also evaluated the Whole-Body Control (WBC) method, a strong traditional baseline, in Isaac. While WBC demonstrates excellent performance in stabilizing body posture, its effectiveness is slightly inferior to that of our proposed approach.

        \subsection{UAV-Quadruped Docking in Simulation}
        As shown in Fig.~\ref{fig:sim_gazebo}, the simulation experiments were conducted in a ROS Gazebo environment running on Ubuntu 20.04. The aerial platform was modeled as the PX4 Iris quadrotor, while the ground platform was represented by the quadruped robot Aliengo. This setup provides a realistic physics-based environment for evaluating the proposed UAV–quadruped docking framework before real-world validation.

\begin{figure}[t]
			\centering
			\includegraphics[width=0.7\linewidth]{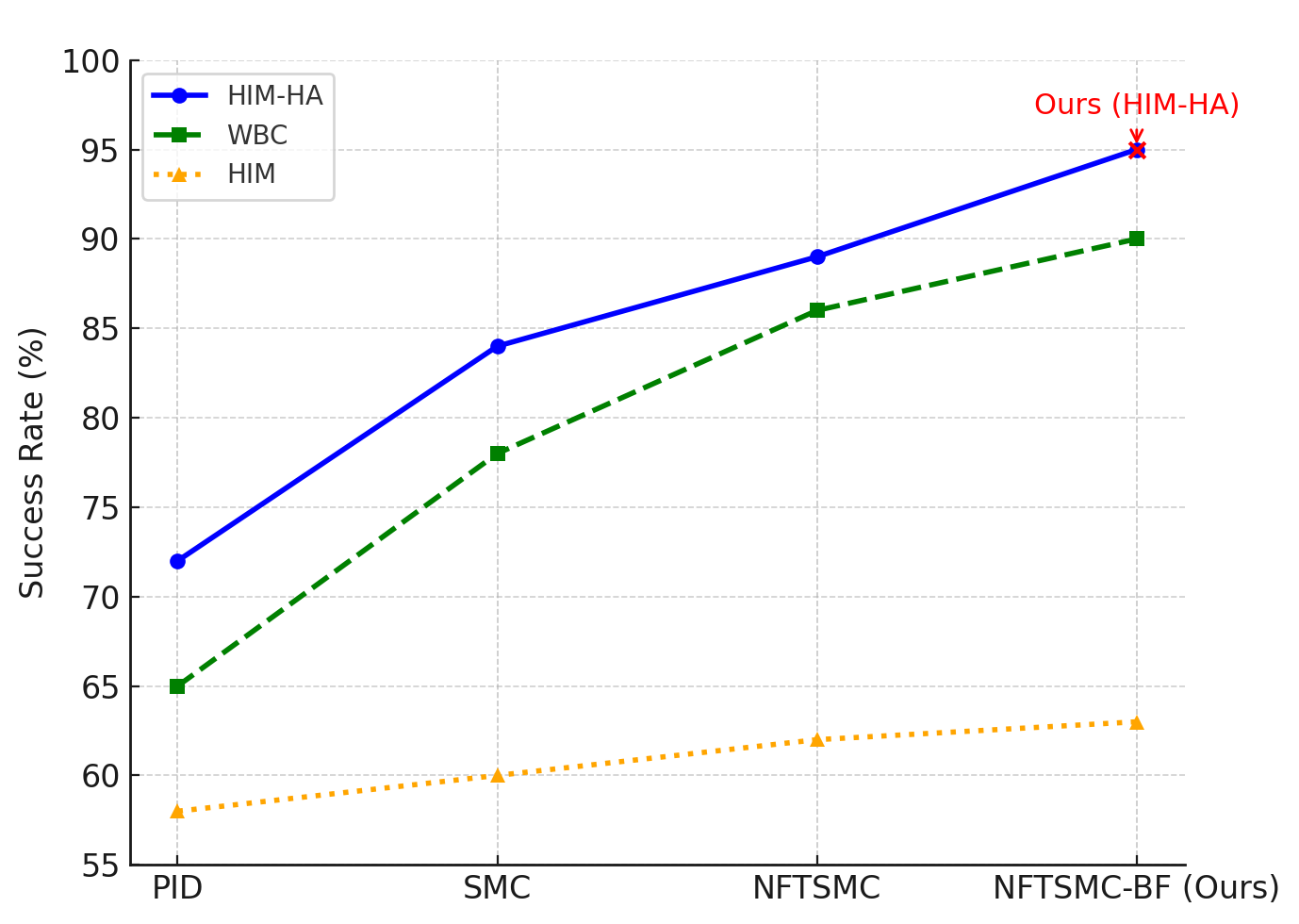} 
			\caption{Docking success rates of PID, SMC, NFTSMC, and the proposed NFTSMC–BF over 100 simulation trials.}
			\label{fig:sim_output} 
		\end{figure}

For the UAV control, we evaluated three baseline algorithms: a conventional PID controller, a standard Sliding Mode Controller (SMC), and an NFTSMC without the Barrier Function (BF). On the quadruped side, our proposed HIM-HA strategy was deployed in Gazebo, with the Whole-Body Control (WBC) method and the original HIM method included as comparisons. The experiments were conducted on a stair scenario with a height of 12 cm. As shown in Fig.~\ref{fig:sim_output}, experimental results demonstrate that the proposed NFTSMC–BF achieves higher docking success rates than the other UAV controllers, while HIM-HA consistently provides a more stable landing platform. It is worth noting that, in principle, the original HIM method without horizontal alignment tends to keep the quadruped body tilted on stairs, making UAV landing difficult; however, in simulation, due to differences in physical parameters from the real robot, successful landings were occasionally still observed.

		\begin{figure*}[t]
			\centering
            \vspace{2mm}
			\includegraphics[width=0.8\linewidth]{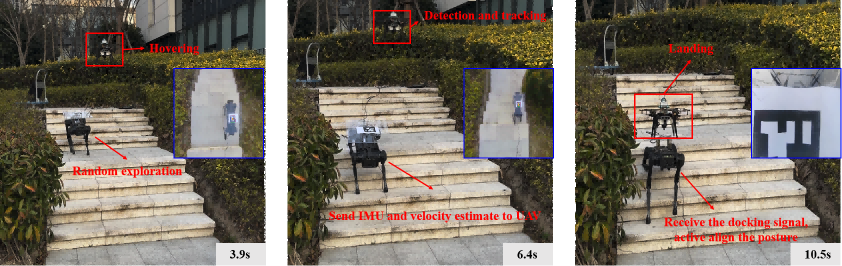} 
			\caption{The docking experiment in an outdoor 17 cm staircase scenario is presented. Two camera perspectives are showcased in this demonstration. The Top view represents footage captured by an external fixed camera, while the Bottom view displays the perspective from the onboard monocular camera located at the underside of the UAV.}
			\label{fig:experiment_1} 
		\end{figure*}

		\subsection{Real-world Experiments}

Fig. \ref{fig:experiment_1} shows the real-world docking process on a 17 cm step-down stair, consisting of three stages. In the acquisition stage, the UAV takes off and hovers until the quadruped enters its field of view (FOV). In the tracking stage, the UAV exploits AprilTag detections, onboard state estimation, and quadruped feedback, while the proposed constraint-aware controller (NFTSMC with BF) ensures accurate close-range tracking under FOV constraints. Finally, the Safety Period (SP) mechanism evaluates tracking accuracy and platform stability. Once satisfied, the UAV descends as the quadruped, controlled by the HIM-HA strategy, aligns its torso horizontally, enabling a safe and stable docking.

	\section{CONCLUSIONS}
	
In this work, we presented an autonomous UAV–quadruped docking framework tailored for GPS-denied environments and complex terrains. The quadruped side employs the HIM-HA strategy to actively stabilize its torso, providing a level landing platform, while the UAV side leverages a constraint-aware controller (NFTSMC with BF) and a Safety Period (SP) mechanism to achieve precise tracking and safe terminal descent.

The proposed framework was validated in real-world experiments, demonstrating successful docking in challenging scenarios such as stairs higher than 17 cm and slopes steeper than $25^{\circ}$. Future work will focus on enabling the quadruped to maintain horizontal alignment while moving faster during docking, and on refining the UAV landing strategy to address multi-axis disturbances caused by rapid quadruped motion in three-dimensional terrains.
	
	


%

%
%
%
%
%

\bibliographystyle{IEEEtran}
\bibliography{output.bib}


\end{document}